\documentclass[11pt,a4paper]{article} % add draft or final to change mode
\usepackage[hyperref]{naaclhlt2019}
\usepackage{times}
\usepackage{latexsym}
\usepackage{url}
\usepackage{graphicx}
\usepackage{footmisc}
\usepackage{amsmath}
\usepackage[utf8]{inputenc}
\usepackage{booktabs}
\usepackage{enumitem}
\usepackage{xargs}
\usepackage[obeyFinal,colorinlistoftodos,textsize=tiny]{todonotes}
\usepackage{xcolor}
\usepackage{soul}
\usepackage{enumitem,amssymb}
\usepackage{pifont}
\usepackage{multirow}
\usepackage{tikz}
\usepackage{verbatim}
\usepackage{amsfonts}
\usetikzlibrary{arrows,shapes,decorations.pathmorphing,automata,backgrounds,petri,positioning,shapes.geometric,decorations.pathreplacing}
\usepackage[belowskip=-5pt,aboveskip=0pt]{caption}
\usepackage{tabularx}

\newcommand{\q}[1]{`#1'}
\newcommand{\dq}[1]{``#1''}

\usepackage{xspace}
\makeatletter
\DeclareRobustCommand\onedot{\futurelet\@let@token\@onedot}
\def\@onedot{\ifx\@let@token.\else.\null\fi\xspace}

\def\eg{\emph{e.g}\onedot} 
\def\ie{\emph{i.e}\onedot} 
\def\cf{\emph{c.f}\onedot} 
\def\etc{\emph{etc}\onedot}

\makeatother

\aclfinalcopy % Uncomment this line for the final submission
\def\aclpaperid{1010} %  Enter the acl Paper ID here
\date{}
\begin{document}
\title{Detecting Cybersecurity Events from Noisy Short Text}
\newcommand*\samethanks[1][\value{footnote}]{\footnotemark[#1]}
\author{
Semih Yagcioglu, Mehmet Saygin Seyfioglu\thanks{\ \  Corresponding author.}, Begum Citamak, Batuhan Bardak\\ \textbf{Seren Guldamlasioglu, Azmi Yuksel, Emin Islam Tatli}\\
STM A.Ş., Ankara, Turkey\\
 {\tt \{syagcioglu, msaygin.seyfioglu, begum.citamak, batuhan.bardak,}\\
 {\tt sguldamlasioglu, azyuksel, emin.tatli\}}
 {\tt @stm.com.tr}}

\maketitle

\begin{abstract}
It is very critical to analyze messages shared over social networks for cyber threat intelligence and cyber-crime prevention. In this study, we propose a method that leverages both domain-specific word embeddings and task-specific features to detect cyber security events from tweets. Our model employs a convolutional neural network (CNN) and a long short-term memory (LSTM) recurrent neural network which takes word level meta-embeddings as inputs and incorporates contextual embeddings to classify noisy short text. We collected a new dataset of cyber security related tweets from Twitter and manually annotated a subset of 2K of them. We experimented with this dataset and concluded that the proposed model outperforms both traditional and neural baselines. The results suggest that our method works well for detecting cyber security events from noisy short text.

\end{abstract}
\section{Introduction}\label{sec:intro}

Twitter has become a medium where people can share and receive timely messages on about anything. People share facts, opinions, broadcast news and communicate with each other through these messages. Due to the low barrier to tweeting, and growth in mobile device usage, tweets might provide valuable information as people often share instantaneous updates such as the breaking news before even being broadcasted in the newswire \cf \citet{petrovic2010streaming}.  %
People also share cyber security events in their tweets such as zero day exploits, ransomwares, data leaks, security breaches, vulnerabilities \etc. %
Automatically detecting such events might have various practical applications such as taking the necessary precautions promptly as well as creating self-awareness as illustrated in Fig. \ref{fig:intro}. Recently, working with the cyber security related text has garnered a lot of interest in both computer security and natural language processing (NLP) communities (\cf \citet{joshi2013extracting,ritter2015weakly,roy2017learning}).
\begin{figure}[t!]
\begin{quotation}
\small{
\noindent Dear @AppleSupport, we noticed a *HUGE* security issue at MacOS High Sierra. Anyone can login as \dq{root} with empty password after clicking on login button several times. Are you aware of it @Apple?
}
\end{quotation}
\caption{\small{A cyber security event. A recently discovered security issue has been reported on Twitter which caught public attention. A security fix has been published right afterward.}} \label{fig:intro}
\end{figure}
Nevertheless, detecting cyber security events from tweets pose a great challenge, as tweets are noisy and often lack sufficient context to discriminate cyber security events due to length limits. %
Recently, deep learning methods have shown to be outperforming traditional approaches in several NLP tasks \cite{chen2014fast,bahdanau2014neural,kim2014convolutional,hermann2015teaching}. Inspired by this progress, our goal is to detect cyber security events in tweets by learning domain-specific word embeddings and task-specific features using neural architectures. 
The key contribution of this work is two folds. First, we propose an end-to-end learning system to effectively detect cyber security events from tweets. Second, we propose a noisy short text dataset with annotated cyber security events for unsupervised and supervised learning tasks. To our best knowledge, this will be the first study that incorporates domain-specific meta-embeddings and contextual embeddings for detecting cyber security events.

\begin{figure*}[!ht]
\centering
\includegraphics[scale=0.55]{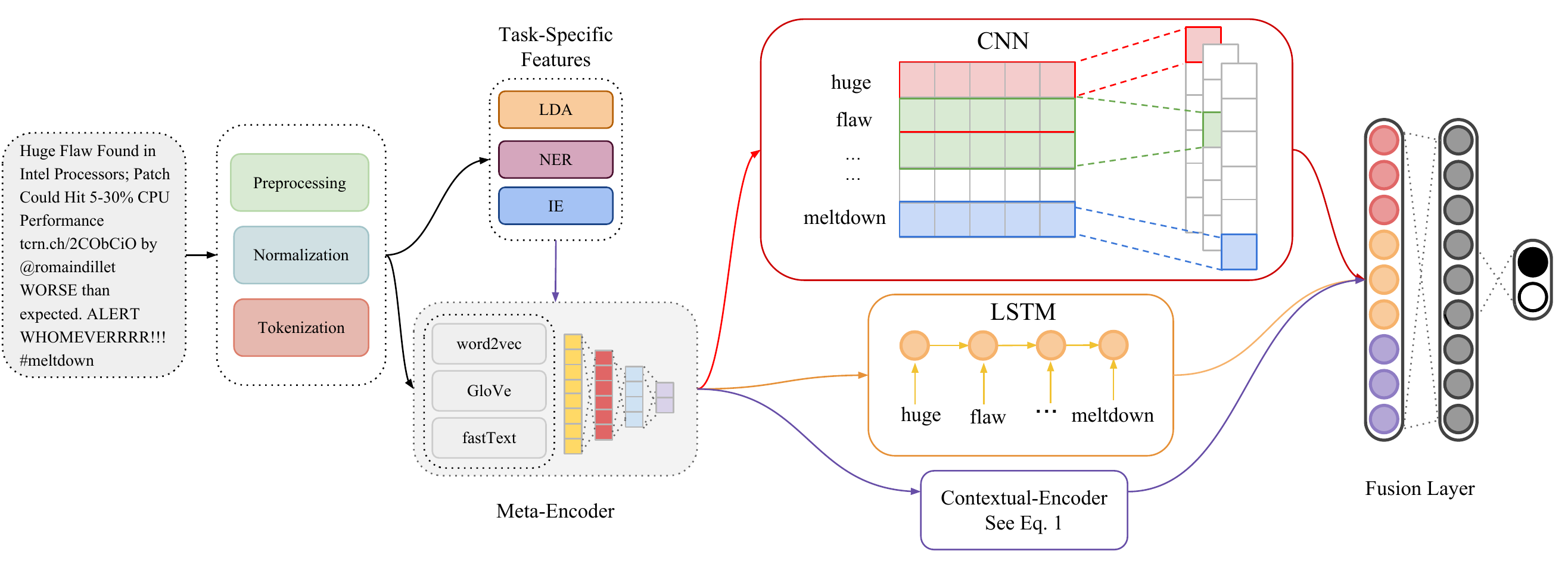}
\caption{\small{System Overview. A tweet is first pre-processed, then task-specific features and word level meta-embeddings are extracted to represent tokens. Finally, a Bi-LSTM, CNN, and Contextual Encoder are fused to classify the encoded tweet.}} 

\label{fig:overview}
\end{figure*}
\section{Method}\label{sec:method}

In the subsequent sections, we address the challenges to solve our task. The proposed system overview is illustrated in Fig. \ref{fig:overview}.

\subsection{Meta-Embeddings}\label{sec:meta}

Word embedding methods might capture different semantic and syntactic features about the same word. To exploit this variety without losing the semantics, we learn meta-embeddings for words.

\noindent\textbf{Word Embeddings.} Word2vec ~\cite{mikolov2013efficient}, GloVe ~\cite{pennington2014glove}, and fastText ~\cite{joulin2016bag,bojanowski2016enriching} are trained for learning domain specific word embeddings on the unlabeled tweet corpus. 

\begin{figure}[b]
\centering
\vspace{-5mm}
\pgfdeclarelayer{bottom}  \pgfdeclarelayer{middle} \pgfdeclarelayer{btop} \pgfdeclarelayer{top} \pgfdeclarelayer{att}
\pgfsetlayers{bottom,middle,btop,top,att}

\begin{tikzpicture}[scale=.65,every node/.style={minimum size=1cm},on grid]
\begin{pgfonlayer}{bottom}

\begin{scope}[  % Upper layer
xshift=-10, yshift=120,every node/.append style={
yslant=0.5,xslant=-1,rotate=-10},yslant=0.5,xslant=-1,rotate=-10
]
\fill[white,fill opacity=.7] (-0.6,0) rectangle (0.6,0.2);
\draw[step=2mm, gray!70] (-.6,0) grid (0.6,0.2);
\draw[black] (-.6,0) rectangle (0.6,0.2);
\draw[violet!20,fill] (0,0) rectangle (0.2,0.2);
\draw[violet!80!black!100] (0,0) rectangle (0.2,0.2);
\draw[step=2mm, violet!70] (0,0) grid (0.2,0.2);

\coordinate(A1) at (0,0);
\coordinate(A2) at (0,0.2);
\coordinate(A3) at (0.2,0.2);
\coordinate(A4) at (0.2,0);
\coordinate(A5) at (0.6,0.1);

\end{scope}
\draw[-latex,thick,opacity=.9] (1.8,4.6) node[right]{{\color{black} \small Meta-Embedding Vector}}to[out=180,in=0] (A5);

\begin{scope}[  % Middle layer 1
yshift=60,every node/.append style={
yslant=0.5,xslant=-1,rotate=-10},yslant=0.5,xslant=-1,rotate=-10
]
\fill[white,fill opacity=.7] (-.6,0) rectangle (1.4,0.8);
\draw[step=2mm, gray!70] (-.6,0) grid (1.4,0.8);
\draw[black] (-.6,0) rectangle (1.4,0.8);

\draw[violet!20,fill] (0,0) rectangle (0.6,0.8);
\draw[violet!80!black!100] (0,0) rectangle (0.6,0.8);
\draw[step=2mm, violet!70] (0,0) grid (0.6,0.8);

\draw[orange!20,fill] (-0.2,0.2) rectangle (0,0.4);
\draw[orange!70!black!100] (-0.2,0.2) rectangle (0,0.4);
\draw[step=2mm, orange!70] (-0.2,0.2) grid (0,0.4);

\coordinate(B1I) at (0,0);
\coordinate(B2I) at (0,0.8);
\coordinate(B3I) at (0.6,0.8);
\coordinate(B4I) at (0.6,0);

\coordinate(B1O) at (-0.2,0.2);
\coordinate(B2O) at (-0.2,0.4);
\coordinate(B3O) at (0,0.4);
\coordinate(B4O) at (0,0.2);
\coordinate (Bs) at (1.5,0);
\coordinate (Bm) at (1.5,0.4);
\coordinate (Be) at (1.5,0.8);

\end{scope}

%\draw[decorate,decoration={brace,mirror, raise=2pt},line width=0.5pt] (Bs) -- (Be) node[right,yshift=2mm,xshift=0mm] {$Z$};

\begin{scope}[	%Middle layer 2
yshift=0,every node/.append style={
yslant=0.5,xslant=-1,rotate=-10},yslant=0.5,xslant=-1,rotate=-10
]
\fill[white,fill opacity=.7] (-.6,0) rectangle (1.8,1.4);
\draw[step=2mm, gray!70] (-0.6,0) grid (1.8,1.4);
\draw[black] (-0.6,0) rectangle (1.8,1.4);

\draw[orange!20,fill] (-0.2,0.0) rectangle (0.4,1.4);
\draw[orange!70!black!100]  (-0.2,0.0) rectangle (0.4,1.4);
\draw[step=2mm, orange!70] (-0.2,0.0) grid (0.4,1.4);

\draw[red!20,fill] (0.8,0.2) rectangle (1,0.4);
\draw[red!70!black!100] (0.8,0.2) rectangle (1,0.4);
\draw[step=2mm, red!70] (0.8,0.2) grid (1,0.4);

\coordinate(C1I) at (-0.2,0.0);
\coordinate(C2I) at (-0.2,1.4);
\coordinate(C3I) at (0.4,1.4);
\coordinate(C4I) at (0.4,0);

\coordinate(C1O) at (0.8,0.2);
\coordinate(C2O) at (0.8,0.4);
\coordinate(C3O) at (1,0.4);
\coordinate(C4O) at (1,0.2);

\coordinate(Cs) at (1.9,0);
\coordinate (Cm) at (1.9, 0.7);
\coordinate(Ce) at (1.9,1.4);

\end{scope}

%\draw[decorate,decoration={brace,mirror, raise=2pt},line width=0.5pt] (Cs) -- (Ce) node[right,yshift=0mm,xshift=3mm] {$K$};

\draw[-latex,thick,opacity=0.9] (3.2,2) node[right]{{\color{black} \small Convolutional Features}}to[out=180,in=0] (Bm);
\draw[-latex,thick,opacity=0.9] (3.2,2) node[right]{{\color{black} \small }}to[out=180,in=0] (Cm);

\begin{scope}[  % Lower layer
xshift=-30, yshift=-50,every node/.append style={
yslant=0.5,xslant=-1,rotate=-10},yslant=0.5,xslant=-1,rotate=-10
]
\fill[white,fill opacity=0.9] (0.4,0) rectangle (3.2,0.6);
\draw[step=2mm, gray!70] (0.4,0) grid (3.2,0.6);
\draw[black] (0.4,0) rectangle (3.2,0.6);
\draw[red!20,fill] (2.0,0) rectangle (2.2,0.6);
\draw[red!80!black!100] (2.0,0) rectangle (2.2,0.6);
\draw[step=2mm, red!70] (2.0,0) grid (2.2,0.6);

%  \draw[blue!20,fill] (1,0) rectangle (1.8,0.2);
% \draw[blue!90] (1,0) rectangle (1.8,0.2);
% \draw[step=2mm, blue!70] (1,0) grid (1.8,0.2); 

\coordinate(D1) at (2.0,0);
\coordinate(D2) at (2.0,0.6);
\coordinate(D3) at (2.2,0.6);
\coordinate(D4) at (2.2,0);
\coordinate (D5) at (3.2,0.3);

\end{scope}

\draw[-latex,thick,opacity=.9] (3.5,0.3) node[right]{{\color{black} \small 3xD Word Embeddings}}to[out=180,in=0] (D5);

\draw[thick,violet!70!black!100,opacity=.7] (A1) -- (B1I);
\draw[thick,violet!70!black!100,opacity=.7] (A2) -- (B2I);
\draw[thick,violet!70!black!100,opacity=.7] (A3) -- (B3I);
\draw[thick,violet!70!black!100,opacity=.7] (A4) -- (B4I);

\draw[thick,orange!70!black!100,opacity=.7] (B1O) -- (C1I);
\draw[thick,orange!70!black!100,opacity=.7] (B2O) -- (C2I);
\draw[thick,orange!70!black!100,opacity=.7] (B3O) -- (C3I);
\draw[thick,orange!70!black!100,opacity=.7] (B4O) -- (C4I);

\draw[thick,red!70!black!100,opacity=.7] (C1O) -- (D1);
\draw[thick,red!70!black!100,opacity=.7] (C2O) -- (D2);
\draw[thick,red!70!black!100,opacity=.7] (C3O) -- (D3);
\draw[thick,red!70!black!100,opacity=.7] (C4O) -- (D4);

\end{pgfonlayer}
\end{tikzpicture}
\vspace{2mm}
\caption{\small{Convolutional encoder as a feature extractor. The decoder is symmetric to the encoder, and in inference time we drop the decoder and use only the encoder network.}}
\label{fig:embedding}
\end{figure}
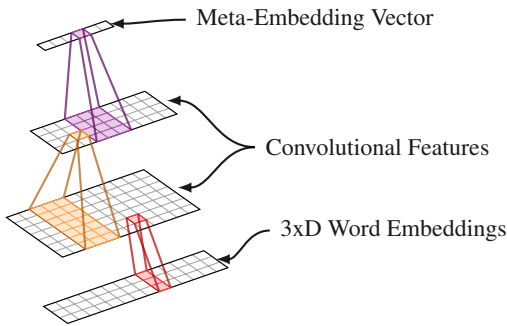

\noindent\textbf{Meta-Encoder.} Inspired by ~\citet{yin2015learning} we learn meta-embeddings for words with the aforementioned word embeddings. We use a Convolutional Autoencoder \cite{masci2011stacked} for encoding $3xD$ size embeddings to a $1xD$ dimensional latent variable and to reconstruct the original embeddings from this latent variable. Both encoder and decoder are comprised of $2$ convolutional layers where $32$ neurons are used on each. The encoder part is shown in Fig. \ref{fig:embedding}.
We argue that this network learns a much simpler mapping while capturing the semantic and syntactic relations from each of these embeddings, thus leading to a richer word-level representation. 
Another advantage of learning meta-embeddings for words is that the proposed architecture alleviates the Out-of-Vocabulary (OOV) embeddings problem, as we still get embeddings from the fastText channel, in contrast to GloVe and word2vec, where no embeddings are available for OOV words.

\subsection{Contextual Embeddings}\label{sec:context}

To capture the contextual information, we learn task-specific features from tweets.

\noindent\textbf{LDA.}\label{sec:summarization}  Latent Dirichlet Allocation (LDA) is a generative probabilistic model to discover topics from a collection of documents \cite{blei2003latent}. LDA works in an unsupervised manner and learns a finite set of categories from a collection, thus represents documents as mixtures of topics. We train an LDA model to summarize each tweet by using the topic with the maximum likelihood \eg with the topic \dq{vulnerability} for the tweet in Fig \ref{fig:intro}.

\noindent\textbf{NER.}
\label{sec:named_entity_recognition}
Named Entity Recognition (NER) tags the specified named entities from raw text into pre-defined categories. Named entities could be more general categories such as people, organizations, or specific entities can be learned by creating a dataset containing specific entity tags. We employ an automatically annotated dataset that contains entities from cyber security domain \cite{bridges2013automatic} to train our Conditional Random Field model using handcrafted features, \textit{i.e.}, uni-gram, bi-gram, and gazetteers. The dataset comprises of ~850K tokens that contain named entities such as \q{Relevant Term}, \q{Operating System},\q{Hardware}, \q{Software}, \q{Vendor}, in the standard IOB-tagging format. Our NER model tags \dq{password} as \q{Relevant Term} and \dq{Apple} as \q{Vendor} for the tweet in Fig \ref{fig:intro}.

\noindent\textbf{IE.} 
Uncovering entities and the relations between those entities is an important task for detecting cyber security events. In order to address this we use Information Extraction (IE), in particular OpenIE annotator\cite{angeli2015leveraging} from the Stanford CoreNLP \cite{manning2014stanford}. Subsequently, we extract relations between noun phrases with the following dependency triplet $\langle arg_1, rel, arg_2\rangle$, where $arg_1$, $arg_2$ denote the arguments and $rel$ represents an implicit semantic relation between those arguments. Hence, the following triplet is extracted from the tweet in Fig.~\ref{fig:intro}, $\langle we,noticed,huge ~security~issue\rangle$.

\noindent\textbf{Contextual-Encoder.} We use the outputs of LDA, NER and IE algorithms to obtain a combined vector representation using meta-embeddings described in Sec.~\ref{sec:meta}. Thus, contextual embeddings are calculated as follows\footnote{We used zero vectors for the non-existent relations.}.\vspace{-2mm}
\begin{equation}
\resizebox{.89\hsize}{!}{$\gamma(\tau) = \dfrac{f(\varphi(\tau)) + \sum\limits_{i=1}^N f(\varsigma(\tau)^i) + \sum\limits_{j=1}^M f(\delta(\tau)^j)}{N + M + 1}$}
\end{equation}
where $\gamma$ function extracts contextual embeddings and $\tau$ denotes a tweet, $f$, $\varphi$, $\varsigma$ and $\delta$ represent meta-embedding, LDA, NER, and IE functions, respectively. Lastly, $N$ and $M$ denote the output tokens.

\subsection{Event Detection}\label{sec:eventdetection}

Inspired by the visual question answering task \cite{antol2015vqa}, where different modalities are combined by CNNs and RNNs, we adopt a similar network architecture for our task. Prior to training, and inference we preprocess, normalize and tokenize each tweet as described in Sec.~\ref{sec:experiments}.

\noindent\textbf{CNN.} We employ a CNN model similar to that of \cite{kim2014convolutional} where we feed the network with static meta-embeddings. Our network is comprised of one convolutional layer with varying filter sizes, that is $2,3,5$. All tweets are zero padded to the maximum tweet length. We use $ReLU$ as activation and global max pooling at the end of CNN.

\noindent\textbf{RNN.} We use a bi-directional LSTM \cite{hochreiter1997long} and read the input in both directions and concatenate forward and backward hidden states to encode the input as a sequence. Our LSTM model is comprised of a single layer and employs $100$ neurons.

\section{Experiments}\label{sec:experiments}
\noindent\textbf{Data Collection.} We collected $2.5M$ tweets using the Twitter's streaming API over a period from \texttt{2015-01-01} to \texttt{2017-12-31} using an initial set of keywords, henceforth referred as \textit{seed keywords} to retrieve cyber security related tweets. In particular, we use the main group names of cyber security taxonomy described in \citet{le2017sonar} as seed keywords e.g.~\q{denial of service}, \q{botnet}, \q{malware}, \q{vulnerability}, \q{phishing}, \q{data breach} to retrieve relevant tweets. Using seed keywords is a practical way to filter out noise considering sparsity of cyber security related tweets in the whole tweet stream. After the initial retrieval, we use \texttt{langid.py} \cite{lui2012langid} to filter out non-English tweets.%

\noindent\textbf{Data Preprocessing.} We substitute user handles with $\$mention\$$, and hyperlinks with $\$url\$$. We remove $emoticons$ and reserved keyword $RT$ which denotes retweets. We substitute hashtags by removing the prefix $\#$ character. We limit characters that repeat more than two times, remove capitalization and tokenize tweets using the Twitter tokenizer in \texttt{nltk} library. We normalize non-standard forms, \ie writing \textit{cu tmrrw} instead of \textit{see you tomorrow}. Although there are several reasons for that, the most prominent one is that people tend to mimic prosodic effects in speech \cite{eisenstein2013bad}. To overcome this, we use lexical normalization, where we substitute OOV tokens with in-Vocabulary (IV) standard forms, \ie a standard form available in a dictionary. In particular we use UniMelb \cite{han2012automatically}, UTDallas \cite{liu2011insertion} datasets. Lastly, we remove identical tweets and check the validity by removing tweets with less than $3$ non-special tokens.

\noindent\textbf{Data Annotation.}~We instructed cyber security domain experts for manual labelling of the dataset. Annotators are asked to provide a binary label for whether there is a cyber security event in the given tweet or not. Annotators are told to skip tweets if they are unsure about their decisions. Finally, we validated annotations by only accepting annotations if at least $3$ among $4$ annotators agreed on. Therefore, we presume the quality of attained ground truth labels is dependable. Overall, $2K$ tweets are annotated.

\noindent\textbf{Dataset Statistics.}~ After preprocessing, our initial $2.5M$ tweet dataset is reduced to $1.7M$ tweets where $2K$ of them are labeled\footnote{Available at \url{https://stm-ai.github.io/}}. The labeled dataset is somewhat balanced as there are $843$ event-related tweets and $1157$ non-event tweets. The training and testing sets have $1600$ and $400$ samples, respectively.

\noindent\textbf{Training.}\label{sec:model_details} We used Keras with Tensorflow backend in our neural models. For fastText and word2vec embeddings we used Gensim, and for GloVe we used \texttt{glove-python} library. For training the word embeddings, we use the entire tweet text corpus and obtain $100$ dimensional word embeddings. We set word2vec and fastText model's alpha parameter to $0.025$ and window size to $5$. For GloVe embedding model, we set the learning rate to $0.01$, alpha to $0.75$ and maximum count parameter to $100$. For embedding models, we determined the minimum count parameter to $5$, culminating in the elimination of infrequent words. Consequently, we have $3$, $100$-dimensional word embedding tensor in which first, second and third channels consist of word2vec, fastText and GloVe embeddings respectively. We then, encode these $3x100$ dimensional embeddings into $1x128$ dimensional representations by using our Meta-Encoder. We train our two channel architecture that combines both LSTM and CNN with $2$ inputs: meta-embeddings and contextual embeddings. We use meta-embeddings for feature learning via LSTM and CNN, and their feature maps are concatenated with contextual embeddings in the Fusion Layer. In the end, fully connected layers and a softmax classifier are added, and the whole network is trained to minimize binary cross entropy loss with a  learning rate of 0.01 by using the Adam optimizer \cite{kingma2014adam}.\footnote{See supplementary for hyperparameter choices.}

\noindent\textbf{Baselines.}~To compare with our results, we implemented the following baselines: \texttt{SVM with BoW}: We trained an SVM classifier using Bag-of-words (BoW) which provides a simplified representation of textual data by calculating the occurrence of words in a document. \texttt{SVM with meta-embeddings}: We trained an SVM classifier with the aforementioned meta-embeddings. \texttt{CNN-Static}: We used \citet{kim2014convolutional}'s approach using word2vec embeddings.

\noindent\textbf{Results.}\label{sec:results} Table \ref{tbl:results} summarizes the overall performance of each method. 
To compare the models, we used four  different metrics: accuracy, recall, precision and F1-score. Each reported result is the mean of a 5-fold cross validation experiment. It is clear that our method outperforms various simple and neural baselines. Also, in Table \ref{tab:resvsgt}, we provide results of our proposed model along with the ground-truth annotations. We also provide results with the different combinations of contextual features, i.e., LDA, NER, IE\footnote{See supplementary for feature combination details.\label{note}}. 

\begin{table}[!t]
\resizebox{\columnwidth}{!}{
\begin{tabular}{lllll}
\hline
\textbf{Models}     & Accuracy & Precision & Recall & F1  \\  \hline
SVM+BoW                     & 0.75     & 0.71      & 0.70   & 0.70        \\ 
SVM+Meta-Emcoder                   & 0.71     & 0.64      & 0.61   & 0.63        \\ 
CNN-static (Yoon Kim, 2014) & 0.76     & 0.72      & 0.69   & 0.70         \\ 
Human                 & 0.65     & 0.70      & 0.87   & 0.59         \\ \hline 

CNN+Meta-Encoder                        & 0.78      & 0.78       & 0.63    & 0.70          \\ 
LSTM+Meta-Encoder                      & 0.78      & 0.74       & 0.70    & 0.72         \\ 
Ours (see Fig. \ref{fig:overview})                        & \textbf{0.82}      & \textbf{0.79}       & \textbf{0.72}    & \textbf{0.76}          \\ 
\hline 
\end{tabular}
}
\centering
\caption{Results}
\label{tbl:results}
\end{table}
\begin{table*}[!t]
\begin{tabularx}{\textwidth}{Xcc}
\hline
\textbf{Tweet}                                                                                                                                                                                     & \textbf{Our Model} & \textbf{GT} \\ \hline
\begin{tabular}[c]{@{}l@{}}that thing where  you run  corporation phishing test and user does'nt\\click it but clicks the next message which is real phishing email sigh\end{tabular} & 0         & 0  \\ \hline
\begin{tabular}[c]{@{}l@{}}march 03 the fbi investigating alleged data breach at the center\\for election systems at kennesaw state university\end{tabular}                             & 1         & 1  \\ \hline
\begin{tabular}[c]{@{}l@{}}cia malware codenames are freaking amazing\end{tabular}                                                                                                                                                & 1         & 0  \\ \hline
\begin{tabular}[c]{@{}l@{}}proprietary software on malware vista10 is more malicious \end{tabular}                                                                                                                                & 0         & 1  \\ \hline
\begin{tabular}[c]{@{}l@{}}in huge breach of trust deidentified medical history data from\\millions of australians can be matched to individuals url2 \end{tabular}                                                                                                                                & 0         & 1  \\ \hline
\begin{tabular}[c]{@{}l@{}}hackers take aim at your $mention$ account with this new phishing\\attack cdwsocia    \end{tabular}                                                                                                                            & 0         & 1  \\ \hline
\begin{tabular}[c]{@{}l@{}}wannacry ransomware ransomwareattack ransomwarewannacry\\malware $url$  \end{tabular}                                                                                                                        & 1         & 0  \\ \hline
\end{tabularx}
\setlength{\tabcolsep}{13em}
\centering
\caption{Some Example Results}
\label{tab:resvsgt}
\end{table*}

\noindent\textbf{Human Study.} $8$ different subjects are thoroughly instructed about what is considered as a cyber security event and individually asked to label $50$ randomly selected tweets from the test set. The results are provided in Table
\ref{tbl:tablehuman}.

\begin{table}[!t]

\resizebox{\columnwidth}{!}{
\begin{tabular}{llllll}
\hline
\textbf{Subjects}                            & Accuracy & Precision & Recall & F1 & Cohen's $\kappa$   \\ \hline
\#1                    & 0.62     & 0.54      & 1   & 0.7  &  \ \ \ \  0.43    \\ 
\#2                   & 0.54    & 0.5      & 0.95   & 0.65   & \ \ \ \ 0.33     \\ 
\#3  & 0.66     & 0.58      & 0.91   & 0.71   &  \ \ \ \ 0.42  \\ 
\#4                 & 0.66     & 0.57      & 1   & 0.73 &  \ \ \ \ 0.46     \\ 
\#5                 & 0.8     & 0.8      & 0.73   & 0.77  &\ \ \ \  0.28     \\ 
\#6                 & 0.66     & 0.57      & 0.95   & 0.72  &\ \ \ \  0.41     \\ 
\#7                 & 0.7     & 0.63      & 0.82   & 0.71  & \ \ \ \ 0.31     \\ 
\#8                 & 0.6     & 0.56      & 0.60   & 0.58  &\ \ \ \  0.28     \\ \hline
\textbf{Average}                 & 0.65     & 0.70      & 0.87   & 0.59  &  \ \ \ \ 0.36     \\ 
\end{tabular}
}
\centering
\caption{Human Study Results}
\label{tbl:tablehuman}
\end{table}

\noindent\textbf{Error Analysis.}
\label{sec:error_analysis} In order to understand how our system performs, we randomly select a set of erroneously classified instances from the test dataset. 
\noindent\textit{Type I Errors.} Our model identifies this tweet as an event \dq{uk warned following breach in air pollution regulation \$url\$} whereas it is clearly about the a breach of a regulation. We hypothesize that this is due to the lack of sufficient training data. Following tweet is also identified as an event \dq{wannacry ransomware ransomwareattack ransomwarewannacry malware \$url\$}. We suspect that the weights of multiple relevant terms deceive the model.

\noindent\textit{Type II Errors.} Our model fails to identify the following positive sample as an event. For \dq{playstation network was the target of miraibotnet ddos attack guiding tech rss news feed search} our model fails to recognize the 'miraibotnet' from the tweet. We suspect this is due to the lack of hashtag decomposition; otherwise, the model could recognize \q{mirai} and \q{botnet} as separate words.

\noindent\textbf{Discussions.}\label{sec:discussions} Cyber security related tweets are complicated and analysing them requires in-depth domain knowledge. Although human subjects are properly instructed, the results of the human study indicate that our task is challenging and humans can hardly discriminate cyber security events amongst cyber security related tweets. To further investigate this, we plan to increase the number of human subjects. One limitation of this study is that we do not consider hyperlinks and user handles which might provide additional information. One particular problem we have not addressed in this work is hashtag decomposition. Error analysis indicates that our model might get confused by challenging examples due to ambiguities and lack of context.

\section{Related Work}
\label{sec:related_work}

Event detection on Twitter is studied extensively in the literature \cite{petrovic2010streaming,sakaki2010earthquake,weng2011event,ritter2012open,yuan2013and,atefeh2015survey}. \citet{banko2007open} proposed a method to extract relational tuples from web corpus without requiring hand labeled data. \citet{ritter2012open} proposed a method for categorizing events in Twitter. \citet{luo2015inferring} suggested an approach to infer binary relations produced by open IE systems. Recently, \citet{ritter2015weakly} introduced the first study to extract event mentions from a raw Twitter stream for event categories DDoS attacks, data breaches, and account hijacking. %
\citet{chang2016expectation} proposed an LSTM based approach which learns tweet level features automatically to extract events from tweet mentions. Lately, \citet{le2017sonar} proposed a model to detect cyber security events in Twitter which uses a taxonomy and a set of seed keywords to retrieve relevant tweets. \citet{tonon2017armatweet} proposed a method to detect events from Twitter by using semantic analysis. \citet{roy2017learning} proposed a method to learn domain-specific word embeddings for sparse cyber security text. Prior art in this direction \cite{ritter2015weakly,chang2016expectation} focuses on extracting events and in particular predicting the events' posterior given the presence of particular words. \citet{le2017sonar,tonon2017armatweet} focus on detecting cyber security events from Twitter. Our work distinguishes from prior studies as we formulate cyber security event detection problem as a classification task and learn meta-embeddings from domain-specific word embeddings while incorporating task-specific features and employing neural architectures.

\section{Conclusion}\label{sec:conclusion}

We introduced a novel neural model that utilizes meta-embeddings learned from domain-specific word embeddings and task-specific features to capture contextual information. We present a unique dataset of cyber security related noisy short text collected from Twitter. The experimental results indicate that the proposed model outperforms the traditional and neural baselines. Possible future research direction might be detecting cyber security related events in different languages.

\section*{Acknowledgments} We would like to thank Merve Nur Yılmaz and Benan Bardak for their invaluable help with the annotation process on this project. This research is fully supported by STM A.Ş. Any opinions, findings and conclusions or recommendations expressed in this material are those of the authors and do not necessarily reflect the view of the sponsor.

\bibliographystyle{acl_natbib}
\bibliography{naaclhlt2018}

\begin{thebibliography}{35}
\expandafter\ifx\csname natexlab\endcsname\relax\def\natexlab#1{#1}\fi

\bibitem[{Angeli et~al.(2015)Angeli, Premkumar, and
  Manning}]{angeli2015leveraging}
Gabor Angeli, Melvin~Johnson Premkumar, and Christopher~D Manning. 2015.
\newblock Leveraging linguistic structure for open domain information
  extraction.
\newblock In \emph{Proceedings of the 53rd Annual Meeting of the ACL 2015}.

\bibitem[{Antol et~al.(2015)Antol, Agrawal, Lu, Mitchell, Batra,
  Lawrence~Zitnick, and Parikh}]{antol2015vqa}
Stanislaw Antol, Aishwarya Agrawal, Jiasen Lu, Margaret Mitchell, Dhruv Batra,
  C~Lawrence~Zitnick, and Devi Parikh. 2015.
\newblock Vqa: Visual question answering.
\newblock In \emph{Proceedings of the IEEE ICCV}, pages 2425--2433.

\bibitem[{Atefeh and Khreich(2015)}]{atefeh2015survey}
Farzindar Atefeh and Wael Khreich. 2015.
\newblock A survey of techniques for event detection in twitter.
\newblock \emph{Computational Intelligence}, 31(1):132--164.

\bibitem[{Bahdanau et~al.(2014)Bahdanau, Cho, and Bengio}]{bahdanau2014neural}
Dzmitry Bahdanau, Kyunghyun Cho, and Yoshua Bengio. 2014.
\newblock Neural machine translation by jointly learning to align and
  translate.
\newblock arXiv:1409.0473.
\newblock Version 7.

\bibitem[{Banko et~al.(2007)Banko, Cafarella, Soderland, Broadhead, and
  Etzioni}]{banko2007open}
Michele Banko, Michael~J Cafarella, Stephen Soderland, Matthew Broadhead, and
  Oren Etzioni. 2007.
\newblock Open information extraction from the web.
\newblock In \emph{IJCAI}, volume~7, pages 2670--2676.

\bibitem[{Blei et~al.(2003)Blei, Ng, and Jordan}]{blei2003latent}
David~M Blei, Andrew~Y Ng, and Michael~I Jordan. 2003.
\newblock Latent dirichlet allocation.
\newblock \emph{JMLR}, 3(Jan):993--1022.

\bibitem[{Bojanowski et~al.(2016)Bojanowski, Grave, Joulin, and
  Mikolov}]{bojanowski2016enriching}
Piotr Bojanowski, Edouard Grave, Armand Joulin, and Tomas Mikolov. 2016.
\newblock Enriching word vectors with subword information.
\newblock arXiv:1607.04606.
\newblock Version 2.

\bibitem[{Bridges et~al.(2013)Bridges, Jones, Iannacone, Testa, and
  Goodall}]{bridges2013automatic}
Robert~A Bridges, Corinne~L Jones, Michael~D Iannacone, Kelly~M Testa, and
  John~R Goodall. 2013.
\newblock Automatic labeling for entity extraction in cyber security.
\newblock arXiv:1308.4941.
\newblock Version 3.

\bibitem[{Chang et~al.(2016)Chang, Teng, and Zhang}]{chang2016expectation}
Ching-Yun Chang, Zhiyang Teng, and Yue Zhang. 2016.
\newblock Expectation-regulated neural model for event mention extraction.
\newblock In \emph{HLT-NAACL}, pages 400--410.

\bibitem[{Chen and Manning(2014)}]{chen2014fast}
Danqi Chen and Christopher Manning. 2014.
\newblock A fast and accurate dependency parser using neural networks.
\newblock In \emph{Proceedings of the 2014 conference on EMNLP}, pages
  740--750.

\bibitem[{Eisenstein(2013)}]{eisenstein2013bad}
Jacob Eisenstein. 2013.
\newblock What to do about bad language on the internet.
\newblock In \emph{HLT-NAACL}, pages 359--369.

\bibitem[{Han et~al.(2012)Han, Cook, and Baldwin}]{han2012automatically}
Bo~Han, Paul Cook, and Timothy Baldwin. 2012.
\newblock Automatically constructing a normalisation dictionary for microblogs.
\newblock In \emph{Proceedings of the 2012 joint conference on EMNLP and
  CoNLL}, pages 421--432. ACL.

\bibitem[{Hermann et~al.(2015)Hermann, Kocisky, Grefenstette, Espeholt, Kay,
  Suleyman, and Blunsom}]{hermann2015teaching}
Karl~Moritz Hermann, Tomas Kocisky, Edward Grefenstette, Lasse Espeholt, Will
  Kay, Mustafa Suleyman, and Phil Blunsom. 2015.
\newblock Teaching machines to read and comprehend.
\newblock In \emph{Advances in NIPS}, pages 1693--1701.

\bibitem[{Hochreiter and Schmidhuber(1997)}]{hochreiter1997long}
Sepp Hochreiter and J{\"u}rgen Schmidhuber. 1997.
\newblock Long short-term memory.
\newblock \emph{Neural computation}, 9(8):1735--1780.

\bibitem[{Joshi et~al.(2013)Joshi, Lal, Finin, and Joshi}]{joshi2013extracting}
Arnav Joshi, Ravendar Lal, Tim Finin, and Anupam Joshi. 2013.
\newblock Extracting cybersecurity related linked data from text.
\newblock In \emph{Semantic Computing (ICSC), 2013 IEEE Seventh International
  Conference on}, pages 252--259. IEEE.

\bibitem[{Joulin et~al.(2016)Joulin, Grave, Bojanowski, and
  Mikolov}]{joulin2016bag}
Armand Joulin, Edouard Grave, Piotr Bojanowski, and Tomas Mikolov. 2016.
\newblock Bag of tricks for efficient text classification.
\newblock arXiv:1607.01759.
\newblock Version 3.

\bibitem[{Kim(2014)}]{kim2014convolutional}
Yoon Kim. 2014.
\newblock Convolutional neural networks for sentence classification.
\newblock arXiv:1408.5882.
\newblock Version 2.

\bibitem[{Kingma and Ba(2014)}]{kingma2014adam}
Diederik Kingma and Jimmy Ba. 2014.
\newblock Adam: A method for stochastic optimization.
\newblock arXiv:1412.6980.
\newblock Version 9.

\bibitem[{Le~Sceller et~al.(2017)Le~Sceller, Karbab, Debbabi, and
  Iqbal}]{le2017sonar}
Quentin Le~Sceller, ElMouatez~Billah Karbab, Mourad Debbabi, and Farkhund
  Iqbal. 2017.
\newblock Sonar: Automatic detection of cyber security events over the twitter
  stream.
\newblock In \emph{Proceedings of the 12th International Conference on
  Availability, Reliability and Security}, page~23. ACM.

\bibitem[{Liu et~al.(2011)Liu, Weng, Wang, and Liu}]{liu2011insertion}
Fei Liu, Fuliang Weng, Bingqing Wang, and Yang Liu. 2011.
\newblock Insertion, deletion, or substitution?: normalizing text messages
  without pre-categorization nor supervision.
\newblock In \emph{Proceedings of the 49th Annual Meeting of the ACL: Human
  Language Technologies: short papers-Volume 2}, pages 71--76. ACL.

\bibitem[{Lui and Baldwin(2012)}]{lui2012langid}
Marco Lui and Timothy Baldwin. 2012.
\newblock langid. py: An off-the-shelf language identification tool.
\newblock In \emph{Proceedings of the ACL 2012 system demonstrations}, pages
  25--30. ACL.

\bibitem[{Luo et~al.(2015)Luo, Luo, and Zhu}]{luo2015inferring}
Kangqi Luo, Xusheng Luo, and Kenny~Qili Zhu. 2015.
\newblock Inferring binary relation schemas for open information extraction.
\newblock In \emph{EMNLP}, pages 555--560.

\bibitem[{Manning et~al.(2014)Manning, Surdeanu, Bauer, Finkel, Bethard, and
  McClosky}]{manning2014stanford}
Christopher~D Manning, Mihai Surdeanu, John Bauer, Jenny~Rose Finkel, Steven
  Bethard, and David McClosky. 2014.
\newblock The stanford corenlp natural language processing toolkit.
\newblock In \emph{ACL (System Demonstrations)}, pages 55--60.

\bibitem[{Masci et~al.(2011)Masci, Meier, Cire{\c{s}}an, and
  Schmidhuber}]{masci2011stacked}
Jonathan Masci, Ueli Meier, Dan Cire{\c{s}}an, and J{\"u}rgen Schmidhuber.
  2011.
\newblock Stacked convolutional auto-encoders for hierarchical feature
  extraction.
\newblock \emph{Artificial Neural Networks and Machine Learning--ICANN 2011},
  pages 52--59.

\bibitem[{Mikolov et~al.(2013)Mikolov, Chen, Corrado, and
  Dean}]{mikolov2013efficient}
Tomas Mikolov, Kai Chen, Greg Corrado, and Jeffrey Dean. 2013.
\newblock Efficient estimation of word representations in vector space.
\newblock arXiv:1301.3781.
\newblock Version 3.

\bibitem[{Pennington et~al.(2014)Pennington, Socher, and
  Manning}]{pennington2014glove}
Jeffrey Pennington, Richard Socher, and Christopher~D. Manning. 2014.
\newblock Glove: Global vectors for word representation.
\newblock \emph{EMNLP}.

\bibitem[{Petrovi{\'c} et~al.(2010)Petrovi{\'c}, Osborne, and
  Lavrenko}]{petrovic2010streaming}
Sa{\v{s}}a Petrovi{\'c}, Miles Osborne, and Victor Lavrenko. 2010.
\newblock Streaming {first} {story} detection with application to twitter.
\newblock In \emph{Human Language Technologies: The 2010 Annual Conference of
  the NAACL}, pages 181--189. ACL.

\bibitem[{Ritter et~al.(2012)Ritter, Etzioni, Clark et~al.}]{ritter2012open}
Alan Ritter, Oren Etzioni, Sam Clark, et~al. 2012.
\newblock Open domain event extraction from twitter.
\newblock In \emph{Proceedings of the 18th ACM SIGKDD international conference
  on KDD}, pages 1104--1112. ACM.

\bibitem[{Ritter et~al.(2015)Ritter, Wright, Casey, and
  Mitchell}]{ritter2015weakly}
Alan Ritter, Evan Wright, William Casey, and Tom Mitchell. 2015.
\newblock Weakly supervised extraction of computer security events from
  twitter.
\newblock In \emph{Proceedings of the 24th International Conference on World
  Wide Web}, pages 896--905. International World Wide Web Conferences Steering
  Committee.

\bibitem[{Roy et~al.(2017)Roy, Park, and Pan}]{roy2017learning}
Arpita Roy, Youngja Park, and SHimei Pan. 2017.
\newblock Learning domain-specific word embeddings from sparse cybersecurity
  texts.
\newblock arXiv:1709.07470.
\newblock Version 1.

\bibitem[{Sakaki et~al.(2010)Sakaki, Okazaki, and
  Matsuo}]{sakaki2010earthquake}
Takeshi Sakaki, Makoto Okazaki, and Yutaka Matsuo. 2010.
\newblock Earthquake shakes twitter users: real-time event detection by social
  sensors.
\newblock In \emph{Proceedings of the 19th international conference on World
  wide web}, pages 851--860. ACM.

\bibitem[{Tonon et~al.(2017)Tonon, Cudr{\'e}-Mauroux, Blarer, Lenders, and
  Motik}]{tonon2017armatweet}
Alberto Tonon, Philippe Cudr{\'e}-Mauroux, Albert Blarer, Vincent Lenders, and
  Boris Motik. 2017.
\newblock Armatweet: Detecting events by semantic tweet analysis.
\newblock In \emph{European Semantic Web Conference}, pages 138--153. Springer.

\bibitem[{Weng and Lee(2011)}]{weng2011event}
Jianshu Weng and Bu-Sung Lee. 2011.
\newblock Event detection in twitter.
\newblock \emph{ICWSM}, 11:401--408.

\bibitem[{Yin and Sch{\"u}tze(2015)}]{yin2015learning}
Wenpeng Yin and Hinrich Sch{\"u}tze. 2015.
\newblock Learning meta-embeddings by using ensembles of embedding sets.
\newblock arXiv:1508.04257.
\newblock Version 2.

\bibitem[{Yuan et~al.(2013)Yuan, Cong, Ma, Sun, and Thalmann}]{yuan2013and}
Quan Yuan, Gao Cong, Zongyang Ma, Aixin Sun, and Nadia~Magnenat Thalmann. 2013.
\newblock Who, where, when and what: discover spatio-temporal topics for
  twitter users.
\newblock In \emph{Proceedings of the 19th ACM SIGKDD international conference
  on KDD}, pages 605--613. ACM.

\end{thebibliography}
\newpage
\clearpage
\section*{Supplementary Notes}
In this supplement, we provide the implementation details that we thought might help to reproduce the results reported in the paper.

\subsection*{What about the model hyperparameters?}

In Table \ref{hyp_parameters}, we provide the hyperparameters we used to report the results in the paper.

\subsection*{Can we download the data?}

Yes. Along with this submission, we provide the whole dataset we collected. Nevertheless, due to the restriction imposed by Twitter, the dataset only contains unique tweet IDs. However, the associated tweets can be easily downloaded with the provided tweet IDs. Dataset is available at \url{https://stm-ai.github.io/}
\begin{table}[b!]

\begin{tabular}{lll}
\hline
 & Hyperparameter & value \\ \hline
general & vector\_size & 100 \\ \hline
\multirow{4}{*}{LDA} & num\_topics & 40 \\
 & update\_every & 1 \\
 & chunksize & 10000 \\
 & passes & 1 \\ \hline
\multirow{4}{*}{w2v \& fastText} & window\_size & 5 \\
 & min\_count & 5 \\
 & iter & 5 \\ 
 & alpha & 0.025 \\ \hline
\multirow{4}{*}{GloVe} & window\_size & 5 \\
 & no\_components & 100 \\
 & learning\_rate & 0.01 \\
 & epoch\_num & 10 \\ \hline
\multirow{4}{*}{Autoencoder} & nb\_epoch & 100 \\
 & batch\_size & 100 \\
 & shuffle & True \\
 & validation\_split & 0.1 \\ \hline
\multirow{2}{*}{CRF} & learning\_rate & 0.01 \\
 & l2 regularization & 1e-2
\end{tabular}
\centering
\caption{Selected Hyperparameters}
\label{hyp_parameters}
\end{table}
\subsection*{How to reproduce the results?}

Here we describe the key steps to recollect data, retrain model and reproduce results on the test set. \begin{itemize}
\item \textbf{Step 1:} As mentioned before, researchers can recollect data through provided tweet IDs.
\item \textbf{Step 2:} After recollecting data, preprocessing, normalization and tokenization tasks are implemented as detailed in Experiments.
\item \textbf{Step 3:} In order to learn domain-specific word embeddings on the unlabeled tweet corpus, meta embedding encoders are trained by applying word2vec, GloVe and fastText as discussed in Section 2.
\item \textbf{Step 4:} Contextual embedding encoder is implemented in order to reveal contextual information as mentioned in Section 2.
\item \textbf{Step 5:} Network architecture combined by CNNs and RNNs is implemented for detecting cyber security related events as detailed in section 2.
\end{itemize}

\subsection*{Have you used a simpler model?}

We favor simple models over complex ones, but for our task, detecting cyber security related events requires tedious effort as well as domain knowledge. In order to capture this domain knowledge, we designed handcrafted features with domain experts to address some of the challenges of our problem. Nevertheless, we also learn to extract features using deep neural networks.

In the Section 3 of the paper, we also provide ablations where we discuss which part of the proposed method adds how much value to the overall success.

\subsection*{Why did you use all of the contextual features?}

At first glance, it might seem that we threw everything that we got to solve the problem. However, we argue that providing contextual features is somewhat yielding a better initialization, thus providing a network to converge better local minima. We also tried out different combinations of contextual features, i.e., LDA, NER, IE by training 2 layered fully connected neural net with them and, although marginally, the combination of all yield the best results, see Table \ref{tab:tablec}. We argue that NER is more biased towards making false positives as it does not consider the word order or semantic meaning and only raises a flag when many relevant terms are apparent. However, results prove that NER's features could be beneficial when used in combination with IE and LDA which indicates that NER is detecting something unique that IE and LDA could not.

\begin{table}[!t]

\begin{tabular}{lll}
\hline
\textbf{Features}       & \textbf{Accuracy}   \\ \hline
All                     & 0.725     \\ 
NER \& LDA                 & 0.705      \\ 
LDA \& IE   			& 0.69      \\ 
NER \& IE              & 0.71         \\ 
IE               	& 0.68      \\ 
NER                	    & 0.64        \\ 
LDA                 	& 0.66       
\end{tabular}
\centering
\caption{Results for Contextual Feature Combinations}
\label{tab:tablec}
\end{table}

\subsection*{How to recollect data?}

As our goal is to develop a system to detect cyber security events, thus collecting more data is crucial for our task. Hence, using the seed keywords as described in the paper Section 3, even more data can be collected using the Twitter's streaming API over a desired period.

\subsection*{What are the most common words?}

Word cloud in Fig. 4 represents the most common words inside the dataset without seed keys. 
\begin{figure}[h]
\centering
\includegraphics[width=\linewidth]{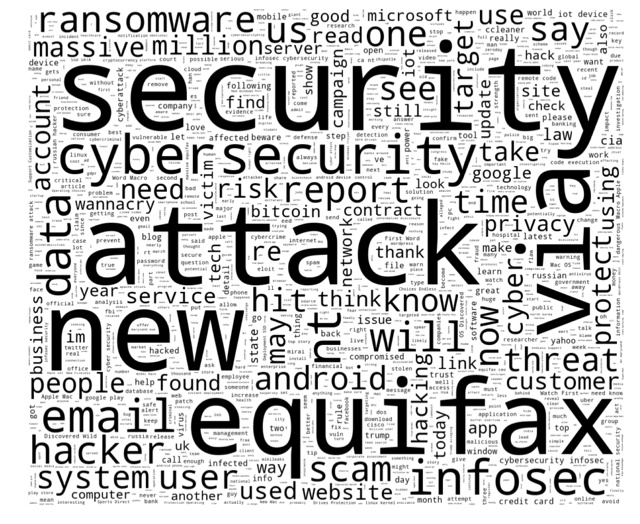}
\caption{\small{Word Cloud}}
\label{fig:confusion}
\end{figure}

\subsection*{How about annotations?}

We expected annotators to discriminate between a cyber security event and non cyber security event. In that regard, we used a team of $8$ annotators, who manually annotated the cyber security related tweets. Each annotator annotated their share of tweets individually, and in sum, the team annotated a total of $2K$ tweets. Following the same procedure, it is possible to annotate more data, which we believe to help achieve even better results. 

\subsection*{How is the human evaluation done?}

We randomly selected  $50$ tweets and provided this subset to $8$ human subjects for evaluation. Each annotator evaluated the tweets independently for his/her share of $50$ tweets. Then, we compared their annotations against ground-truth annotations.

\subsection*{What about hardware details?}
All computations are done on a system with the following specifications: NVIDIA Tesla K$80$ GPU with $24$ GB of VRAM, $378$ GB of RAM and Intel Xeon E$5$ $2683$ processor.

\end{document}